\documentclass{cta-author}
{}
{}
{}
\usepackage{geometry}
\usepackage{mathptmx}
\usepackage{wrapfig}
\usepackage{hyperref}
\usepackage{biblatex}
\fontfamily{pbk}\selectfont
\usepackage{mathtools}
\usepackage{graphicx,color}
\hypersetup{
  colorlinks,
  citecolor=blue,
  linkcolor=blue,
  urlcolor=blue}

\begin{document}


\title{An advanced combination of semi-supervised Normalizing Flow \& Yolo (YoloNF) to detect and recognize vehicle license plates.}

\author{\au{Khalid OUBLAL$^{1\corr}$}, 
\au{Xinyi DAI $^{ 2}$}}

\address{\add{1}{Corresponding author: Department of Computer Science, Ecole Polytechnique - IP Paris}
\add{2}{Department of Computer Science, Ecole Polytechnique - IP Paris}\email{khalid.oublal@polytechnique.edu}}

\begin{abstract}
Fully Automatic License Plate Recognition (ALPR) has
been a frequent research topic due to several practical applications. However, many of the current solutions are still not robust enough in real situations, commonly depending on many constraints. This paper presents a robust and efficient ALPR system based on the state-of-the-art YOLO object detector and Normalizing flows. The model uses two new strategies:
Firstly, a two-stage network using YOLO and a normalization flow-based model for normalization to detect Licenses Plates (LP) and recognize the LP with numbers and Arabic characters. Secondly, Multi-scale image transformations are implemented to provide a solution to the problem of the YOLO cropped LP detection including significant background noise. Furthermore, extensive experiments are led on a new dataset with realistic scenarios, we introduce a larger public annotated dataset collected from Moroccan plates. We demonstrate that our proposed model can learn on a small number of samples free of single or multiple characters. The dataset will also be made publicly available to encourage further studies and research on plate detection and recognition.\\

\end{abstract}
\maketitle
\section{Overview}
\label{sec1}
Intelligent transportation is one of the main areas where artificial intelligence has gained importance. Automatic License Plate Recognition (ALPR) is an important task in smart transportation and surveillance, which has many practical applications such as automatic traffic enforcement, such as the detection of stolen vehicles or toll violations, traffic flow control, etc.\par
The ALPR problem can be divided into three sub-tasks: license plate detection, license plate segmentation, and character recognition. These subtasks constitute the common pipeline of ALPR systems found in the literature \cite{saudi_Realtime,yolotiny,pakistani}, and many works focus on only one or two of these subtasks \cite{saudi,openALPR,ALPRmar, OCR}. One of our main contributions is to introduce a new combination of normalizing flows and YOLO as object detectors which supersedes the segmentation and character recognition part, making the task becomes a simple multi-scale labeling problem.\par

Our work is based on a normalizing flow model, proposed in 2020 by \citeauthor{same}  \cite{same}, this model uses a latent space of normalizing flow representing a normal sample's feature distribution. Unlike other generative models such as variational autoencoder (VAE) \cite{VAE} and GANs \cite{Gans}, the flow based generator makes the bijective mapping between feature space and latent space assigned to a likelihood.\par

Normalizing Flows (NF) are a network that can generate
complex distributions by transforming a probability density
from a \textbf{latent space} through a series of invertible affine transformations of <<flows>> and was introduced by \citeauthor{karami2019invertible}\cite{karami2019invertible}. Based on the change of variable rule, the bijective mappings between feature space and latent space can be evaluated in both directions. The formula is derived as Eq.\ref{equation_derive}. Thus a scoring function can be derived to decide if an image contains an anomaly or not. As a result, most common samples will have a high likelihood, while uncommon images will have a lower likelihood.\par
\begin{equation}
p_i(\textbf{z}_i) = p_{i-1}(f_i^{-1}(\textbf{z}_i)) \left\vert\det\dfrac{d f_i^{-1}}{d \textbf{z}_i} \right\vert\label{equation_derive}
\end{equation}

\begin{equation*}
\text{where } \textbf{z}_i = f_{i}(\textbf{z}_{i-1}) \textbf{, thus }\mathbf{z}_{i-1} = f_i^{-1}(\textbf{z}_i) \sim p_{i-1}(\textbf{z}_{i-1})
\end{equation*}
This model aims to find the best probability distribution
$P_{z}(z)$ in the latent space Z to maximize likelihoods for
extracted features X. According to the change-of-variables
formula extended to the function \ref{equation_derive}, the following steps are illustrated by the equation (Eq.\ref{equation_derive_finale}). After adding log function on both sides, the loss
function can be defined as (Eq.\ref{equation_loss}).

\begin{equation}
\begin{split}
p_i(\textbf{z}_i) 
&= p_{i-1}(f_i^{-1}(\textbf{z}_i)) \left\vert \det\dfrac{d f_i^{-1}}{d \textbf{z}_i} \right\vert\\
&= p_{i-1}(\textbf{z}_{i-1}) \left\vert \det {\Big(\dfrac{d f_i}{d\textbf{z}_{i-1}}\Big)^{-1}} \right\vert \\
&= p_{i-1}(\textbf{z}_{i-1}) {\left\vert \det \dfrac{d f_i}{d\textbf{z}_{i-1}} \right\vert^{-1}}\\
\log p_i(\textbf{z}_i) &= \log p_{i-1}(\textbf{z}_{i-1}) - \log \left\vert \det \dfrac{d f_i}{d\textbf{z}_{i-1}} \right\vert
\end{split}
\label{equation_derive_finale} 
\end{equation}

\begin{equation}
\label{equation_loss}
\log p_i(\textbf{z}_i) = \log p_{i-1}(\textbf{z}_{i-1}) - \log \left\vert \det \dfrac{d f_i}{d\textbf{z}_{i-1}} \right\vert
\end{equation}

\begin{figure}[h!]
    \centering
    \includegraphics[scale=0.37]{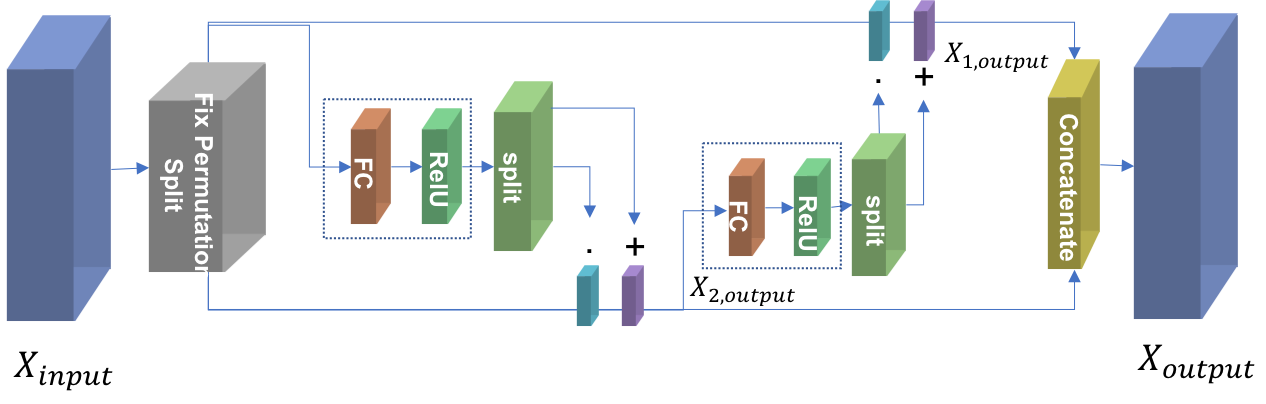}
    \caption{Composition schemes for affine coupling layers proposed by \citeauthor{coupling_layers}\cite{coupling_layers}.\label{coupling}}
\end{figure}

YOLO or You Only Look Once Unified and real-time object detection presented in 2016 by \citeauthor{redmon2016you}\cite{redmon2016you} reframes the object detection as a regression problem that spatially separates bounding boxes and associates their class probabilities. A single convolutional neural network is used to predict bounding boxes and class probabilities in the YOLO system. With a given image as input, the system first divides the image into an S x S grid. Each cell predicts B bounding boxes and their corresponding confidence scores. The confidence score is defined as $Pr(Class_i)\times IOU^{truth}_{b-box}$, where the intersection on the union (IOU) between the ground truth and the predicted bounding box is calculated. Then, the conditional class probabilities are multiplied by the confidence scores of each bounding box to obtain confidence scores for a specific class. In this paper, we are interested in this YOLO structure and we will detail all the main elements to combine this structure with the other Normalizing Flows model structure which uses a coupling layer architecture as proposed in Real-NVP \cite{coupling_layers}. The layout of a single block is given in Figure \ref{coupling}.\par

In this paper, we focus on the YOLO neural network and we will detail all the main elements to combine this network with the Normalizing Flows network. For this purpose, this paper is organized as follows: firstly we briefly review related work in Section.\ref{sec2}. Then, the Moroccan ALPR dataset with the new database is presented in Section.\ref{sec3}. Section.\ref{sec4} presents the novel combination of YOLO and flow normalization which builds the proposed ALPR system. We present and discuss the results of our experiments in section.\ref{sec5}. Conclusions and future work are presented in Section.\ref{sec6}.

\section{Related works}\label{sec2}
Multiple ALPR methods have been tested recently on detection and recognition of LPs, however the use of deep learning as a tool for ALPR targetting Moroccan LPs has not been sufficiently explored. In \cite{ALPRmar}, authors have attempted a hybrid model based on morphological extraction and use an optical character recognition OCR phase as \cite{saudi_Realtime} which is based on Tesseract \cite{OCR}.\par
For this purpose, we briefly review several recent works that use Deep Learning approaches in the context of ALPR very similar to Moroccan licenses plates. Specifically, we discuss work related to each stage of ALPR.

\subsection{License Plates detection}
End-to-end deep learning approaches for license plate recognition include \cite{pakistani,saudi}, which detect proposed license plate regions and perform the final selection using a CNN as a binary classifier. As proposed by \citeauthor{malays}, the CNN is trained for the entire character sequence to detect and recognize LPs with ConvNet-RNN netowrk first\cite{cheang2017segmentation}.\par Using such techniques, it is impossible to separate license plates from other general alphanumeric text, such as when the vehicle image is taken in a severe weather scenario or a simple area that is not centered on the image while other text of the same shape, such as an advertisement, is in the scope of primary view. In one instance, the fixed width of the bounding box limits the application to standard license plates. \cite{pakistani} applied a visual attention model to detect license plates containing blue and yellow regions of interest. The eventual classification was completed by a CNN model that mapped the offset using the Support Vector Machine (SVM) model. The results were not as expected because the SVM model significantly limits some LP types and makes the method more sensitive to illumination and scale variations. Another solution proposed in \cite{jain2016deep} which consisted in generating the license plates with a vertical Sobel filter and applying a binary CNN for the final verification. However, the results of this approach failed to handle the actual negative cases caused by Sobel due to the high noise on the plate images. \par
\citeauthor{yolotiny} proposed another method to meet the specific needs of Moroccan license plates using the small version of YOLO, this method present a relevant results in terms of detection but it remains very sensitive to the variation of class type since it is trained directly on regular Moroccan plates while it fails on the plates of trucks or service vehicles since the position of the LPs is not on the large ground despite the fact that the content is the same. Moreover when a several of vehicles appear in the same frame the network does not detect all the plates and the ALPR interested only on the one which present in the large ground, this is due probably to the Tiny-YOLOv3 Layers which reduces the number of convolutional layers so the features are extracted only by using a small number of 1×1 and 3×3 convolutional layers.

\subsection{Licenses Plates Recognition \& Segmentation}
For the recognition process, the studies we consulted often use networks that can convert a text-image into editable text. For the recognition process, we used an attention sequence network, which takes the rectified image as input and predicts the sequence of characters of the license plate. Their network consists of an encoder and a decoder. The encoder uses a convolutional layer stack (ConvNet) that extracts the characters to improve detection against other test approaches.\\
\citeauthor{menotti2014vehicle}\cite{menotti2014vehicle}proposed using random CNNs to extract features. Using random CNNs to extract features for character recognition achieved much better performance than using image pixels or learning filter weights by backpropagation. On the other hand, \cite{chang2004automatic} proposed to perform china characters recognition only as a sequence of Multi-scale labeling problem, which is not far from the approach of \cite{yolotiny} that also renders it as a character-level classification problem.  A recurrent neural network (RNN) with connectionist temporal classification (CTC) is used to label the sequence data, thus recognizing the entire LP without character-level segmentation. The latter, however, is sensitive to variance, especially since it is a problem of labeling on images correlated with the plates, so very small for which the neural network type RCNN\cite{RCNN} or Fast RCNN\cite{girshick2015fast}, YOLO-tiny easily fails as a result of missing of more features as well as the gathering of Arabic characters types diverges significantly  this problem.
\subsection{Remarks}
To summarize our remarks, these multiple methods cited their experience on Moroccan LP datasets that present both Arabic and Latin alpha-numeric characters. Making it difficult to accurately evaluate the methods presented. Moreover, the one using Yolo-tiny \cite{yolotiny} presents large difficulties, which make the ALPR system fails in the global classification of characters, this is due to the sensitivity of the model to the noise since this detection and recognition is performed only on one-stage way which increases the chances of losing the confidence rate. In addition to the uses of small CNN penalize the detection of small objects, such as the characters. This problem can be improved by using a normalizing flow approach if we adapt the density estimation using REAL NVP approach proposed by \citeauthor{dinh2016density}\cite{dinh2016density} to our case.\par
To that effect we use an improved YOLO-v3 object detection at each step \cite{tian2019apple} to create a robust and efficient end-to-end ALPR system. In addition, we augment and transform the data for character recognition, as this step is the main problem of some ALPR systems including that of \cite{yolotiny}.

\section{The ALPR dataset}\label{sec3}

\begin{figure}[h!]
    \centering
    \includegraphics[scale=0.20]{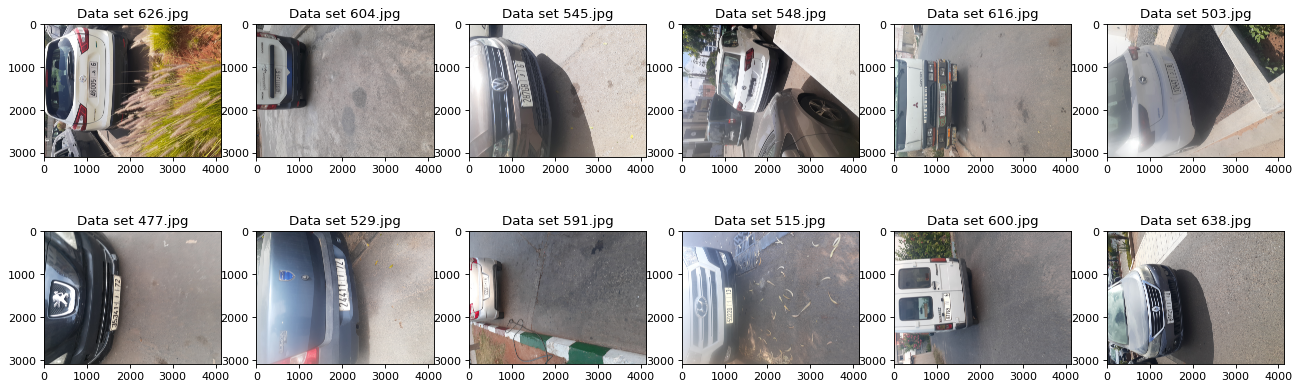}
    \caption{Samples from Dataset of vehicles}
    \label{LPs}
\end{figure}

The public ALPR Moroccan dataset contains 657 images taken inside a vehicle in irregular traffic or parking at the urban environment. We also add to this, some samples from the public dataset of UPM6 \cite{UPM6} as well as the HDR dataset dataset LPs \cite{HDRdataset} which is not far from the characteristics of the Moroccan LPs for the new cars only (present of \textbf{ww, M} characters) and the Arabic characters extracted from the Arabian open source database in order to perform the characteristics engineering step with consistency. We have more than $1,200$ images and more than $1800$ images of sample characters generated by cropping from the Moroccan dataset and other from HDR dataset and generated images following the proposed method of \cite{GANs_images} using GAN models .
Approximately 85\% cars, 10\% motorcycles with new and old plates and approximately 5\% trucks. In Morocco, the LPs have a slight variation in size and variations in color (black, white, red) depending on the type of vehicle and its category e.g. official or public LP service cars are often black and red. In general,LP cars have a size of 40cm $\times$ 13cm, while motorcycles have 20cm $\times$ 17cm . Figure.\ref{LPs} shows some of the different types of LPs in the dataset.

\begin{figure}[h!]
    \centering
    \includegraphics[scale=0.20]{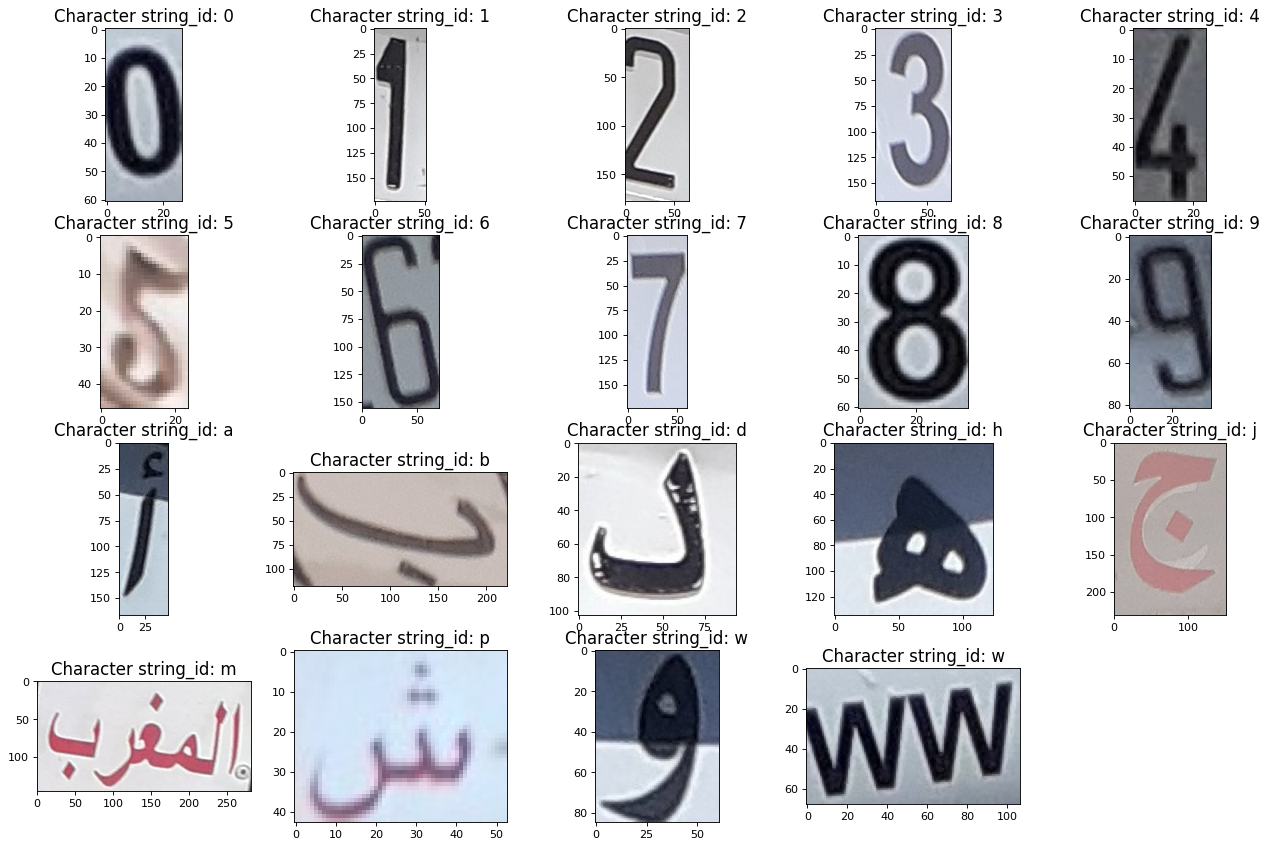}
    \caption{Samples from Dataset of characters}
    \label{characters}
\end{figure}
The data set is divided as follows: 40\% for training, 40\% for testing and 20\% for validation.
The distribution of the dataset was done in such a way that each split has the same number of images obtained, taking into account the type and position of the vehicle, the color and characters of the PL vehicle's . In order to facilitate the interpretation, we associate to each character a Latin label and for the numbers the numbers themselves, and if there is a Latin character we took it itself.
\begin{figure}[htbp]
    \centering
    \includegraphics[scale=0.44]{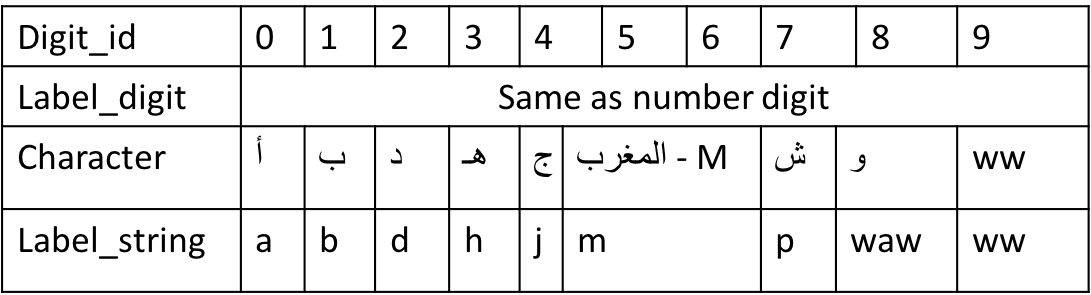}
    \caption{Latin labels attributed to each class of the dataset characters (Figure.\ref{characters}).}
    \label{label_characters}
\end{figure}
In Morocco, LPs is different from license plates which use Latin letters such as the European ones. There is a small similarity between Moroccan plates and some that use Arabic or Persian letters such as Algeria, Arabic Saudi, Emirate..or Pakistanis plates.\\
Moroccan LPs consist of 3 parts separated by two verticals lines, in the first part on the left there are five numbers that identify the vehicle $XXXXX\in\{0\to99999\}$, in the middle an Arabic characters and on the right one or two numbers $YY \in \{0\to99\}$ that represent the prefecture or the province where the vehicle has been registered: For example, 38 for <<Ouarzazate>> province, the plate can be typed as $XXXXX-a-38$ or for the capital province <<Rabat>> $XXXXX-a-1$. \par
\begin{figure}[htbp]
    \centering
    \includegraphics[scale=0.20]{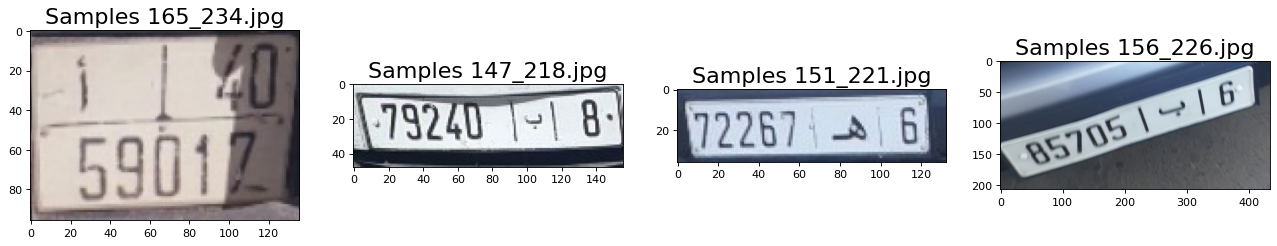}
    \caption{Illustration of the different images License plates and their appearance ( the right image begin with Arabic character while the other doesn't, and the image of the motorcycle has no character Arabic, and has only numbers.}
    \label{different_plates}
\end{figure}

The rule that we have just explained is not satisfied across the entire dataset, there are some older vehicles with plates that start with Arabic characters, then the number that refers to the province of registration and then the vehicle number, this example is shown in Figure \ref{different_plates} .\par

\begin{figure}[htbp]
    \centering
    \includegraphics[scale=0.60]{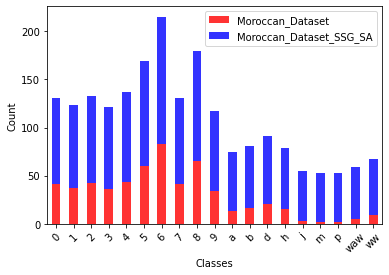}
    \caption{Characters distribution in the new Morocco-ALPR dataset Enhanced by other examples from the HDR LPs datasets\cite{HDRdataset} of the same type, compared to the one provided to the challenge \cite{MoroccoAI} and \cite{NARSA}.}
    \label{hist_string}
\end{figure}

 Therefore, the digits numbers \{5,6,8\} and the characters \{a,b,d,h\} have many more samples than the others, as shown in Figure.\ref{hist_string} which refer this problem of ALPR to \textbf{unbalanced} data which must be taken into account during the implementation of our approach.

\section{Proposed ALPR approach}\label{sec4}
\begin{figure}[h!]
    \centering
    \includegraphics[scale=0.4]{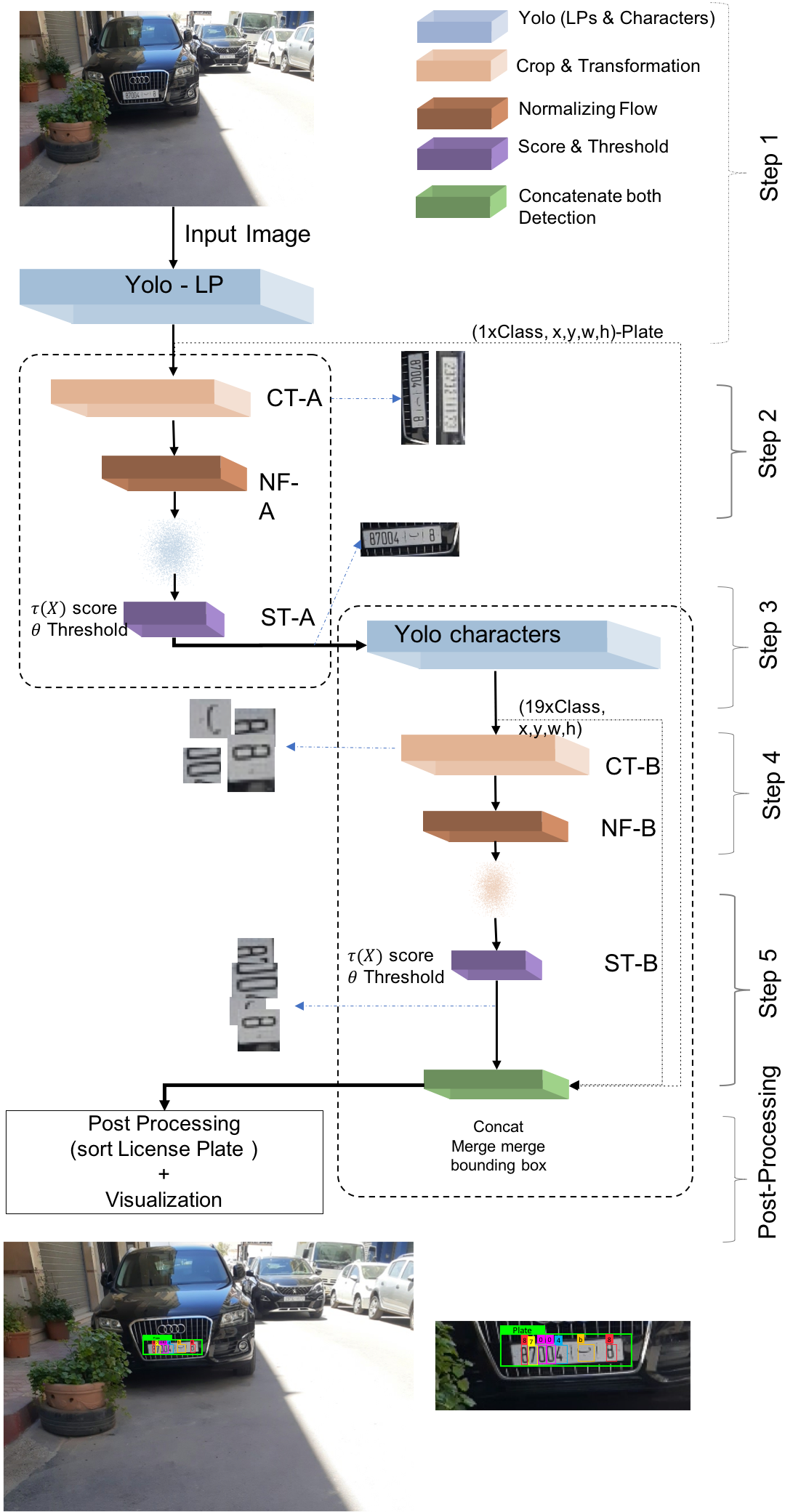}
    \caption{Global Proposed ALPR Approach (YoloFN)\label{proposed_approach}}
    \label{fig:config_1_2}
\end{figure}

The proposed approach is based on 5 subsections: Plate detection, distribution verification of the detected LPs, then we move to characters detection and again to the cross-checking and linking two types of classification results of the detected LPs and finally to character recognition. For visualization model we also introduced to plot predicted bounding boxes for each plate and the predicted characters result on each image (Figure.\ref{proposed_approach}).

\begin{enumerate}
\item The proposed model takes vehicle images as input and then using YOLOv3 \textbf{(Yolo-LP)}, trained on our custom dataset, to detect and draw bounding boxes on each plate.\\

\item An extraction and Transformation block \textbf{(CT-A)} is created to crop the plate images based on the bounding boxes drawn by the preview step. Both of the cropped plates images and the original images will be saved. The transformation model is further introduced to rotate and crop each plates image’s edges to remove the background noises surrounding the plate.\\ \label{step3}

\item The processed plates images are then passed into the
normalizing flow model \textbf{(NF-A)} to generate a normal distribution by maximum likelihood training only on plate (e.g semi-supervised training). A scoring function is used to compute likelihood function to classify the input samples as plates or not this step improves the precision of the detection. The unit \textbf{(ST-A)} allows to keep only the detection with the largest confidence in cases where more
than one Plate is detected by introducing threshold value $\theta$. We have found that the most optimal value of $\theta$ is $0.1$. \\

\item A second block of Yolo-v3 \textbf{(Yolo Characters)} which trained only on character dataset (this is not available from the ANRT, for that we have created own labelling ) to detect and draw bounding boxes on each characters included numbers. The second block \textbf{(CT-B)} provide provide samples of cropped and transformed characters; \\

\item The block \textbf{(NF-B)}, will decide if a character is present on the distribution built by the Normalization-Flow trained only on the 19 classes of characters, The \textbf{(CT-B)} will decide according to the score result if there is a real correlation between the character and the ones seen during the training step, this is performed by the scoring function $\theta$ if this score is much lower, this means that the character corresponds to a class, this one is the class detected with the greatest probability, this promotes the use of the \textbf{Softmax} function to retrieve the most likely class for a character combine with the score results of Normalizing flow. \\

The robustness of this approach is not only due to the fact that the Flow Normalizing can check the class detected by linking it to the scoring function, but rather the possibility of transforming the samples avoid errors in detection and classification, which by default is based only on confidence. 

\end{enumerate}

\subsection{Detect vehicles Plates by Yolo-v3 and Normalizing Flow}
\subsubsection{Block Yolo-V3}\label{yolo}
In order to use YOLO to detect LP for the first stage or character in the fourth stage, we need to change the number of filters in the last convolutional layer to match the number of classes. YOLO uses A anchor boxes to predict bounding boxes, in our case we proposed $A = 5$ with four coordinates $(x, y, w, h)$, confidence and C class probabilities, so the number of filters is given by:
\begin{equation}
    FILTERS = (C + 5) \times A
\end{equation}

For the second time we also thought to mix two models, one that detects 2 classes (vehicle, motorcycles), which is already trained on coco dataset. For training YOLOv3 and Fast-YoLo we used pretrained weights on ImageNet \cite{yolov3}, and another that only detects the plate, but most of our images the full vehicle structure is not taken, which makes this idea useless. Therefore we will have only one class which is the Licenses Plates, below the description of filters=$(1 + 5)\times5=30$ that we used.
\begin{figure}[htbp]
    \centering
    \includegraphics[scale=0.5]{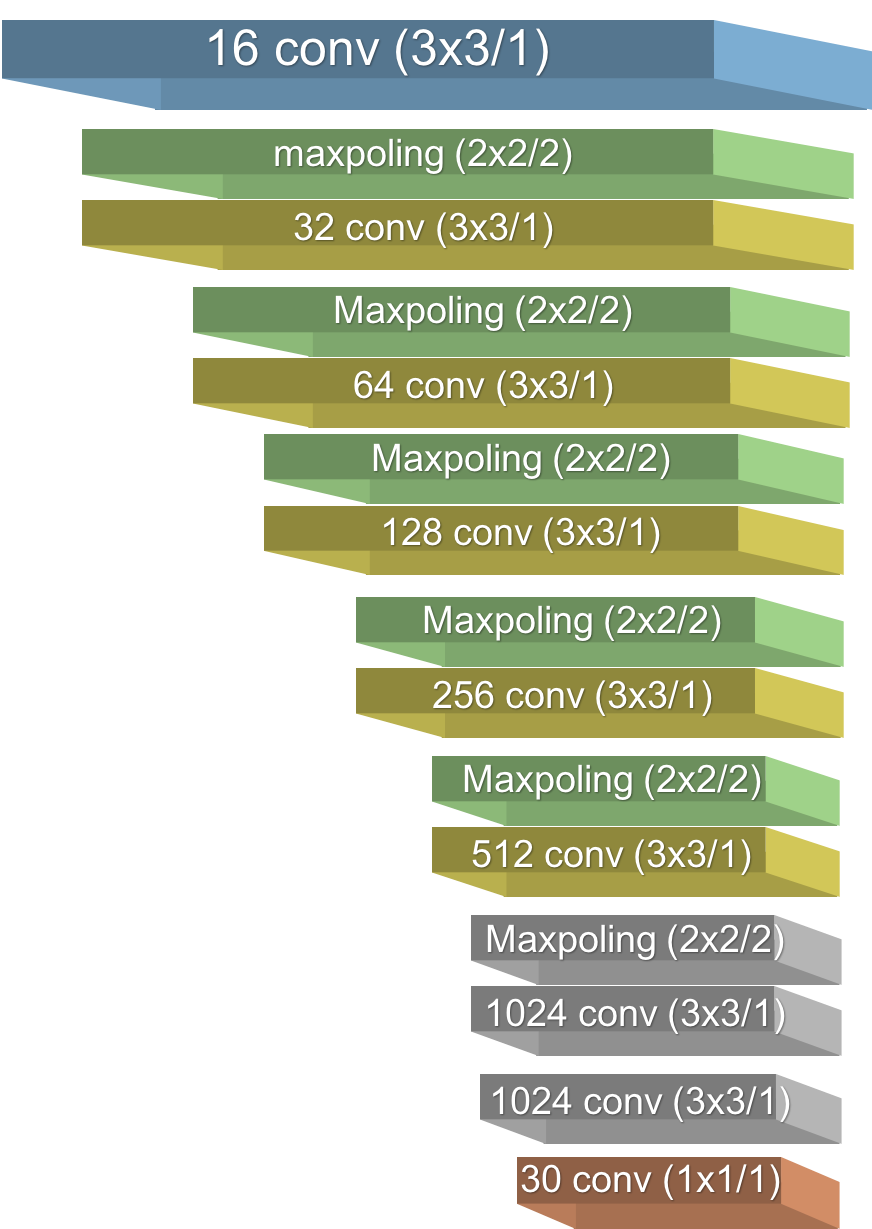}
    \caption{The network used for LP detection. There are 30 filters in the last layer to detect a single class.}
    \label{fig:my_label}
\end{figure}

\subsubsection{Block Normalizing Flow \& Transformation}
We used model Different \cite{same} which is a state-of-art model that utilizes a latent space of normalizing flow to represent normal samples’ feature distribution, which is in our case LPs, and for the second stage of our pipeline, features are digits, Arabic characters.
To improve the performance of the model on the output images from Yolo, we propose an image transformation
model to rotate and crop each LP image. In training, various scales of cropping on LPs images are performed to ease
background noise interference. Moreover, the range of rotation
for input images is reduced from 180 degrees to 30 or 20
degrees for better computing performance. Then the transformed images are fed into a pre-trained AlexNet\cite{ding2018alexnet} to extract the feature. The extracted feature map is
further passed into a normalizing flow-based on coupling layer (Figure \ref{coupling}) to output a normal distribution of LPs by maximum likelihood training.

As \citeauthor{same}\cite{same} proposed, we used the calculated likelihoods as a criterion to classify a sample as an authentic plate (respectively character) if it fits with the distribution generated from the learned distributors or it is not. To get a robust score $\tau(x)$, we average the negative $log-likelihoods$ using multiple transformations $T_{i}(X)\in T$ of an image $X$:
\begin{equation}
    \tau(X) = E_{T_{i \in T}}(- \log p_{Z}(f_{NF}(f_{ex}(T_{k}(X)))))
\end{equation}

Here, we have not provided for any adjustments,\cite{same} has T chosen rotations and manipulations of brightness and contrast (which is consistent with our color image dataset and is sensitive to variations in contrast and brightness since most images are taken during the day). An image is classified as LP respectively characters if the score $\tau(X)$ is greater than the threshold value $\theta$.

To classify if an input image is Licenses Plates (or character in the case of Normalizing Flow Characters) or not, we will use the \cite{same} structure which uses a scoring function that compute the average of the \textbf{negative log-likelihoods} using multiple transformations $T_k(X)$ of an image $X$.
\begin{equation}
L(\mathcal{D}) = -\frac{1}{\vert\mathcal{D}\vert} \sum_{\mathbf{X} \in \mathcal{D}} \log p(\mathbf{X})
\end{equation}

\begin{figure}[htbp]
    \centering
    \includegraphics[scale=0.3]{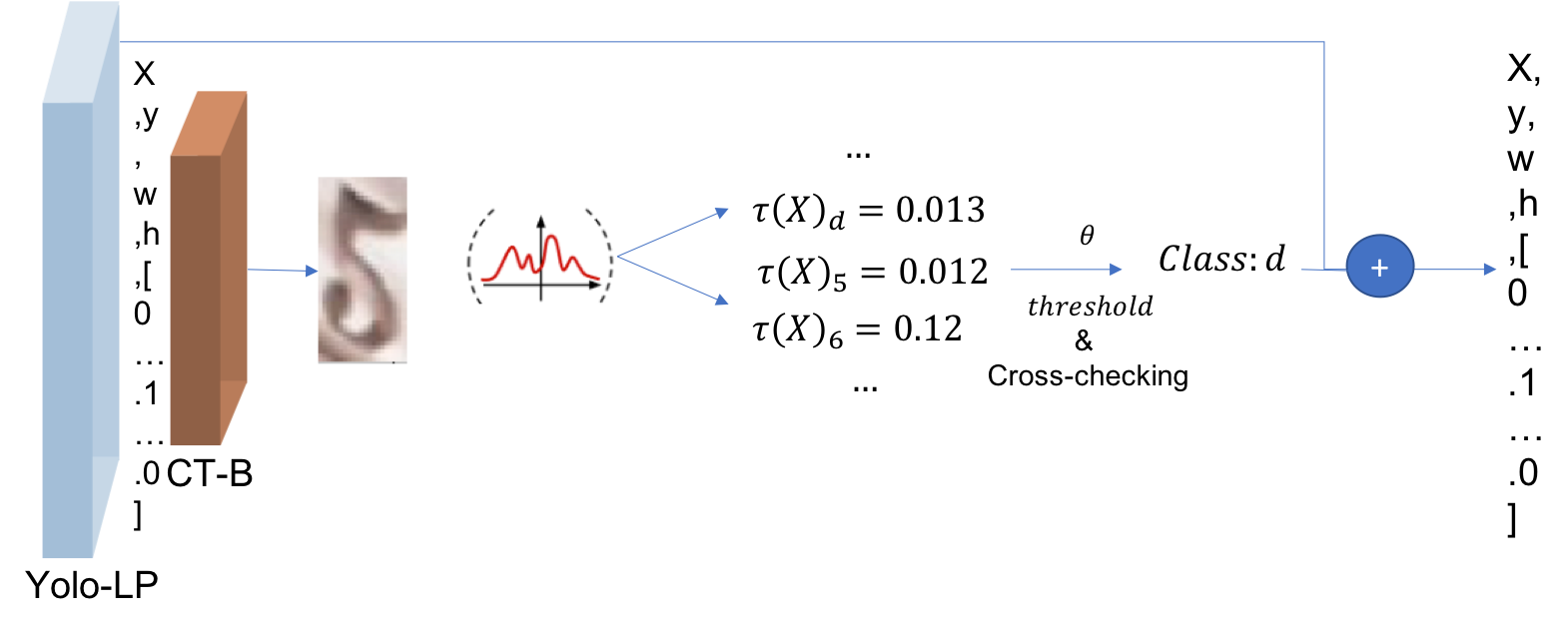}
    \caption{A simple illustration of Flow Normalizing on a samples}
    \label{data}
\end{figure}

\subsection{Detection and Recognition of Characters }
For the detection and recognition of characters, we built
a new network inspired on the YOLOv3 architecture (e.g as the first stage), with
fundamental technical differences to accommodate 19 classes: 10 numerical digits, 5 Arabic characters,
($0,\to,9$) and (a, b, d, h, j, m, p, waw, ww), consequently, we have to define
filters=(19+5)x5=120. Please
refer to Section \ref{yolo} and Table.\ref{tab1}.\\

\begin{table}[htbp]
\processtable{The input image is a
240 x 80 color LP patches, and the output of 30 x 10 guaranteed
enough horizontal granularity for the 19 characters (digits and Arabic characters).\label{tab1}}
{\begin{tabular*}{20pc}{@{\extracolsep{\fill}}lllll@{}}\toprule
Layer &  Filters & Size & Input &  Output\\
\midrule
conv & 32  & 3 × 3 / 1 & 240 × 80 × 3  & 240 × 80 × 32\\
max  &   & 2 ×  2 / 2  & 240 × 80 × 32 & 120 × 40 × 32\\
conv & 64  & 3 × 3 / 1 & 120 × 40 × 32 & 120 × 40 × 64\\
max  &    & 2 × 2 / 2  & 120 × 40 × 64 & 60 × 20 × 64\\
conv & 128 & 3 × 3 / 1 & 60 × 20 × 64  & 60 × 20 × 128\\
conv &  64  & 1 × 1 / 1 & 60 × 20 × 128 & 60 × 20 × 64\\
conv & 128 & 3 × 3 / 1 & 60 × 20 × 64  & 60 × 20 × 128\\
max  &    & 2 × 2 / 2 & 60 × 20 × 128 & 30 × 10 × 128\\
conv & 256 & 3 × 3 / 1 & 30 × 10 × 128 & 30 × 10 × 256\\
conv & 128 & 1 × 1 / 1 & 30 × 10 × 256 & 30 × 10 × 128\\
conv & 256 & 3 × 3 / 1 & 30 × 10 × 128 & 30 × 10 × 256\\
conv & 512 & 3 × 3 / 1 & 30 × 10 × 256 & 30 × 10 × 512\\
conv & 256 & 1 × 1 / 1 & 30 × 10 × 512 & 30 × 10 × 256\\
conv & 512 & 3 × 3 / 1 & 30 × 10 × 256 & 30 × 10 × 512\\
conv & 120  & 1 × 1 / 1 & 30 × 10 × 512 & 30 × 10 × 80\\
detection& & & \\
\botrule
\end{tabular*}}{}
\end{table}

Finally, we perform a post-processing step in order to swap
highly confused letters to digits and digits to letters, since
the Brazilian license plate is formed by exactly three letters
followed by four digits.

As we mentioned before, a Moroccan LPs is different from license plates which use Latin letters such as European. There is a small similarity between Moroccan plates and some that use Arabic or Farsi letters such as Algeria, Arabic Saudi, Emirate or Pakistanis LPs which we have refereed on section \ref{sec2}.

Since our evaluation only takes the new version of the name (XXXXXX-character-XX) where X is digit number ($0\to9$), there is an important post-processing step in order to sort the plates correctly.

Understanding the specific layout of the Moroccan plate, we sort the detected characters by their horizontal position for the vehicles. The left (four to six characters) of the Arabic letter is the vehicle number ($0\to99999$) as format XXXXX, then the right of it is the prefecture/province number ($0\to 99$) as XX. Otherwise, if the left side is empty, then the number of the prefecture/province is on the right side and all characters at the bottom of the Arabic character correspond to the number of vehicles. When there is no Arabic character, the sorting list is defined as default (from left to right and from top to bottom).

\section{Experiments \& results}\label{sec5}

All experiments were performed on an NVIDIA Titan XP GPU (3,840 CUDA cores and 32 GB of RAM) using the Darknet \cite{darknet13} framework and different \cite{same} implemented on PyTorch.

We consider as correct only detection with IoU $\geq$ 0.5. This value was chosen after several trials to improve the performance of our model on the validation dataset. In addition, the following parameters were used for training the networks: 80k iterations (max batches) and learning rate$=[10^{-3}, 10^{-4}, 10{-5}]$ with batch size of 64 and 1000 iterations.

\subsection{Evaluation metric}
\begin{itemize}
    \item \textbf{mAP \& loss Likelihood metric}: As we have well specified concerning the Normalizing Flow network, its evaluation metric consists mainly on the likelihood function, so to evaluate the training of Yolo in the two steps of our ALPR system, we have considered Recall \& precision curve which allows us to have an access to the IOU and the threshold value that we must apply to increase the precision.\\
    
    The Mean Average Precision (mAP) is more accurate than F1-scores because it takes into account the precision and recall relationship in a holistic manner. The intuitive use of this metric is related to the official metric used in \cite{cheang2017segmentation}.
    \item \textbf{Intersection over Union (IoU)}: To train an object detection model, usually, there are 2 inputs:An image and ground-truth bounding boxes for each object in the image. The model will predicts the bounding boxes of the detected objects. It is expected that the predicted box will not match exactly the ground-truth box
        \begin{equation}
        IoU=\frac{\text{Intersection area}}{\text{Union area}}
    \end{equation}
    \item {\bf Recall \& Precision}
    \begin{equation}
    \text{Recall} = \frac{TP}{TP+FN} \text{ , } \hspace{1cm}
    \text{Precision} = \frac{TP}{TP+FP}
    \label{equation_rappel}
    \end{equation}
    
    \item {\bf mAP formula}:\\
    \begin{equation}
        mAP=\frac{1}{N}\sum_{k=1}^{k=N}AP_{k}
    \end{equation}
    Where the $AP_{k}$ is the Average precision of class $k$ and $N$ is the number of classes.
    \item \textbf{Levenshtein distance metric:} Our last evaluation is mainly related to the string similarity between string LP detected and the annotated label on the validation dataset. In order to achieve this, we use Levenshtein distance like Hamming distance, is the smallest number of edit operations required to transform one string into the other. Unlike Hamming distance, the set of edit operations also includes insertions and deletions, thus allowing us to compare strings of different lengths. This metric appears robust and easy to implement.
    In table  we show the average error created by the proposed method compared to other methods given in the state-of-the-art of the report.
\end{itemize}

\subsection{Training and hyperparameters}

Since we have a large number of trainable parameters, we must use techniques to reduce \textbf{overfitting}. First, The Darknet model used an exponential decay of the learning rate that we set to our desired values, to ensure that the change in parameters will be small over time. In addition, we used \textbf{L1} and \textbf{L2} regularization in the dense layers, as they disperse the error terms across all weights, leading to more accurate final models. To achieve this, the model was based on DarkNet-53. Third, we used an early stop callback to stop the training when the validation no longer improves based on the validation loss.

The training model converges after $80$ iterations and is stable for the rest of the training phase. The average validation loss is $0.21$.

\begin{figure}[h!]
    \centering
    \includegraphics[scale=0.60]{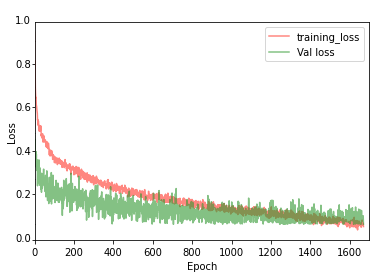}
    \caption{Training and validation loss of Yolo-LP (Improved version of Yolo-v3).}
    \label{training_loss_vall_yolo_LP}
\end{figure}

\begin{figure}[h!]
    \centering
    \includegraphics[scale=0.61]{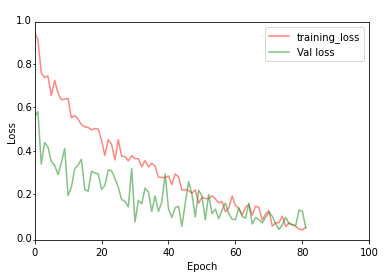}
    \caption{Training and validation loss of Yolo-Characters (Improved version of Yolo-v3).}
    \label{training_loss_vall_yolo_characters}
\end{figure}

As the dataset we are using an imbalanced data some classes have a really multiples samples (around of 20\%) while more than 90\% of the other classes have a relatively low samples. To avoid this issue we using the label smoothing which convert our one-hot encoded labels to a smooth probability distribution \cite{zhang2019bag}.
\begin{figure}[htbp]
    \centering
    \includegraphics[scale=0.52]{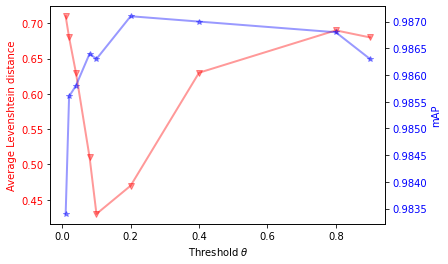}
    \caption{The mAP and Average Levenshtein distance as a function of theta threshold values tested to find the optimal value.}
    \label{threshold}
\end{figure}\\

As well as training with the characters provided in the training set, we also augment the data in two ways. Firstly, we create negative images to simulate characters from other vehicle categories, and then we also verify which characters can be rotated horizontally and vertically to create new cases.

\subsection{Evaluation on the ALPR Dataset}

For LP detection, we find that in the most challenging images, e.g., motorcycle images, the vehicle's LP is not entirely within the predicted bounding box, requiring a margin change of about 5\% in the validation set for the entire LP to be entirely within the predicted bounding box of the vehicle. 
Therefore, we use a margin of 15\% in the test set and in the training of the LP detection Yolov3.
The first stage, i.e. the YOLO model dedicated to LP detection \textbf{Yolo-LP} at a recall rate obtained is 0.9813 which correspond to 1178/1,200 images. We were not able to detect the LP in a single vehicle in only 22 images, because a false positive was predicted with a higher confidence+score function $\tau (image)$ than the real LP.\\

Therefore, we used the \textbf{Yolo-characters} results to perform post-processing in cases where more than one LP is detected, for example: evaluate on each detected LP if there are at least $4$ characters or consider only the LP where the confidence of the characters is higher and the likelihood scoring function is minimal (e.g $\tau(X)$). However, since the actual LP can be detected with very low confidence levels (i.e., $\leq$ 0.2 to 0.1), some false negatives should be analyzed. To understand what the network is seeing we used GradCAM\cite{selvaraju2016grad} implemented at the output of Yolo and block Normalizing Flow.\\

\begin{figure}[htbp]
    \centering
    \includegraphics[scale=0.60]{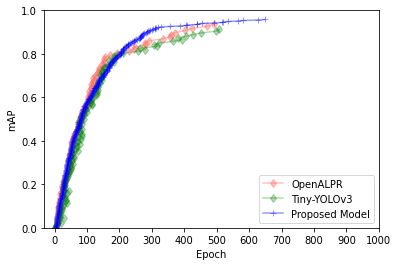}
    \caption{Proposed System mAP and training batch curve in comparison to other methods Tiniy-Yolo \cite{yolotiny} and OpenALPR used on Saudi vehicles \cite{openALPR}}
    \label{mAP}
\end{figure}

\begin{table}[htbp]
\processtable{Results obtained and the Levenshtein distance score in each ALPR stage in the Moroccan APLR dataset (Validation set). The Recall stands for detection and segmentation, and mAP stands for recognition.\label{tab2}}
{\begin{tabular*}{15pc}{@{\extracolsep{\fill}}llll@{}}\toprule
 ALPR &  & Recall/mAP  & Average Levenshtein\\
 &  &   &distance\\
\midrule
& Vehicle Detection.       & 1 &    \\
& License Plate Detection  & \textbf{0.9871}  &   \\
Tiny& Character Detection   & 0.9675   &  0.9654  \\
Yolo & Character Recognition    & 0.9683 &    \\

\midrule
& Vehicle Detection.       & -  &    \\
& License Plate Detection  & 0.98  &   \\
Open& Character Detection(Seg)   & 94.75  & 1.4654    \\
ALPR& Character Recognition    & 92.83  &    \\
\midrule
& Vehicle Detection.       & 1  &    \\
YoloFN& License Plate Detection  & 0.9813  &   \\
(our& Character Detection   & \textbf{0.9863}  &    \textbf{0.4654} \\
proposed)& Character Recognition    & \textbf{0.9871} &    \\

\botrule
\end{tabular*}}{}
\end{table}
\begin{figure}[htbp]
    \centering
    \includegraphics[scale=0.23]{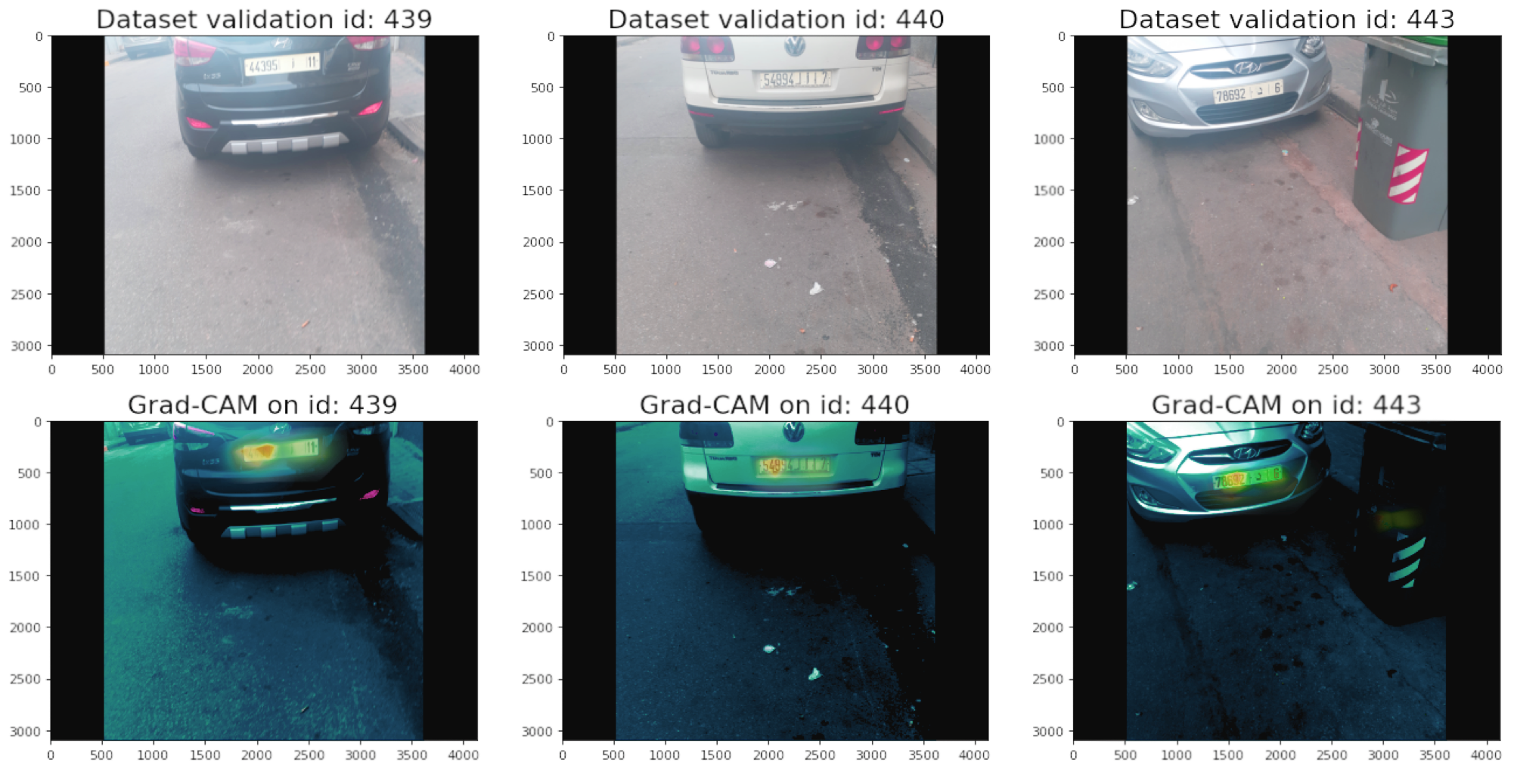}
    \caption{Result of the Grad-Cam on license plate detection, on the validation set id:439, 440, and 443.}
    \label{Grad_CAM}
\end{figure}
\begin{figure}[htbp]
    \centering
    \includegraphics[scale=0.50]{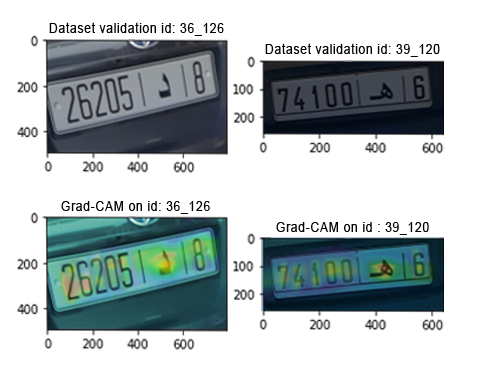}
    \caption{Result of the Grad-Cam on license plate detection, on the validation set id:36\_126 and 39\_120}
    \label{Grad_CAM_2}
\end{figure}
For the character detection and classification part (stage 2) In the validation set, a margin of 20\% is required for each cropped detection of LP in order to contain all its characters. Although this margin adds background noise in the LPs patches, the model is robust to this noise because at the output of block flow (\textbf{NF-B}, Figure.\ref{proposed_approach}) normalizing with a threshold $\theta$ of 0.1 manages to detect the characters cleanly. The recall obtained is 0.9863 and 0.9854 considering the whole dataset test. This is much higher than the \cite{yolotiny} method, which is explained by the fact that block Flow normalizing cross-checking class by class and the sum of the scores evokes a low error on the classification. Moreover, the transformation at block (\textbf{CT-B}, Figure.\ref{proposed_approach}) allows to better check the class of a given character, especially the similarity between the class \textbf{\{h\}} and \textbf{\{d\}} in some cases as well as the class \textbf{\{1\}} and \textbf{\{a\}}. Figure.\ref{mAP} illustrates the mean average when training the openALPR and the Yolo-Tiny models for character detection.
Our proposed model obtain a better results in the validation set of APLR Morocco dataset, our model achieved good score of recall, mAP, and also minimize the average Levenshtein distance (ALD) compared to other methods, this but we note that the detection of the licenses at the level of stage 1 there is a similarity between Yolo-tiny with a slight advantage of the latter, but this is explained by our choice close to a theta that reduces ALD, however the detection of the characters, our model is significantly better than openAPLR and Yolotiny.

Figure.\ref{data} (Appendix \ref{appendix3}) shows some LPs of different categories correctly detected despite the tilt of the plate or the weather that degrades the quality of the images.

The table \ref{tab2}, gives the average error of this distance on which the APLR system provides at the output a string of predict characters were it's compared to target label. Note that for our YoloFN model we changed several values of theta at the threshold part \textbf{ST-B} of YoloFN $\theta\in[0.6 to 0.01]$ to achieve an optimal score of ALD (Figure.\ref{threshold}), and the corresponding value of theta is $\theta=0.1$ this value was chosen to calculate the error on the final ALD on test dataset.

\section{Conclusions \& future works}\label{sec6}

In this paper, we introduce a new combination in cascade between two neural networks as well as a new dataset that contain the Moroccan license plates. This dataset presents several challenges related to types of LPs, background noise and the environment where the image was taken. Moreover, the method proposed a detection of the LPs and then cross checking if it is the class LP and not something else, then the network detects again the characters and the recognition of the class is done by combining the confidence and the error on the distribution with the normalizing flow model. The result showed that the model increases more than 2.3\% of its mAP and reduces more than 3\% of the errors on the predicted string of the LP if compared with a YOLO without the normalizing flow blocks. We also compared our results on the same validation set with the OpenAPLR method that uses OCR and the recognition improvement exceeds 7.1\%.\\

Several improvements of this model are also possible despite the chosen value of $\theta$ (theta), which corresponds to the minimum of the Levenshtein distance, this value does not correlate to the maximum mAP which means that we have lost more than 1.2\% of mAP to satisfy the criterion of Average Levenshtein distance. Adding more layers to the bock flow normalization could solve this particular issue and find a most efficient value of $\theta$.  Finally, we conclude that based on these extensive experiments this proposed approach is robust for the detection of LPs and it can be extended to a real time implementation with an FPS around $30Fps$ (using GPU).\\

This work can be further extended to tackle the detection of small overlapping objects in real time in more challenging contexts. For example, the proposed solution can be further tuned to detect pedestrians in challenging scenes subject to extreme conditions such as storms and snow. It can be used in search and rescue operations to save lives.

\section{Acknowledgments}\label{sec11}

This work was supported by a prize from the National Telecommunications Regulatory Agency (ANRT) and the Nvidia Deep Learning Institute. The author would like to thank MoroccoAI conference data challenge team for the coordination and organization.

\textit{
<<Please note that this research study is a result of an ALPR framework that we propose as a solution to the problem of LP detection and recognition for vehicles in Morocco (Data challenge conference of December 2021), we have tried to detail as much as possible each step, however due to constraints related to the submission deadline we have not been able to explore in depth some parts including the implementation of Grad-Cam on a YOlo, this will be published as soon as possible. However, I am available on \href{mailto:khalid.oublal@polytechnique.edu}{khalid.oublal@polytechnique.edu} to answer for any question related to this framework. I would like to thank the MoroccoAI conference organizing team, the ANRT and Nvidia for this real challenge.>>}


\printbibliography

\setcounter{section}{0}
\section{APPENDIX A: Annotation of new dataset Moroccan ALPR-Plus}\label{appendix1}
We note that a new dataset \cite{alprmaroc} of annotated images has been introduced in order to increase the dataset of the Moroccan ALPR, moreover this dataset contains several levels of labeling (vehicles, plates care, and characters) and is being performed using VOCA format. We have injected some images from HDR dataset\cite{HDRdataset} and others was generated by using LP-GAN model proposed by \citeauthor{GANs_images} \cite{GANs_images}. 
\begin{itemize}
\item The final injection conditions: \begin{itemize}
            \item If all the characters in the plate are numbers or digits with a character included in the list of classes available on the original Moroccan ALPR dataset.
            \item Respect the order and the colors of most of the plates by adding only white (or black plates written in gray or white). \end{itemize}
\end{itemize}
\section{APPENDIX C: Examples of LPs detection and recognition success and failures cases}\label{appendix3}





\begin{center}
\centering
\begin{figure}[htbp]
    \centering
    \includegraphics[scale=0.45]{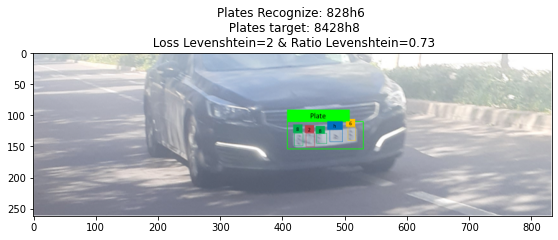}
    \caption{Plates Recognize: 828h6, Plates target: 8428h8, Levenshtein distance=2 \& Ratio Levenshtein=0.73}
    \label{Nh2}
\end{figure}

\begin{figure}[htbp]
    \centering
    \includegraphics[scale=0.45]{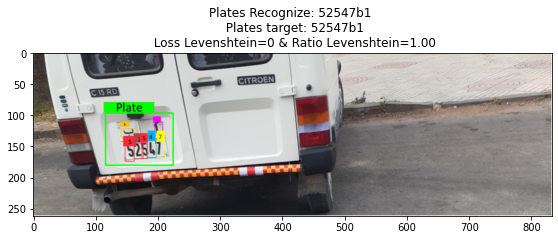}
    \caption{Plates Recognize: 52547b1, Plates target: 52547b1, Levenshtein distance=0 \& Ratio Levenshtein=1.00}
    \label{fig_Nh2}
\end{figure}

\begin{figure}[htbp]
   \includegraphics[scale=0.45]{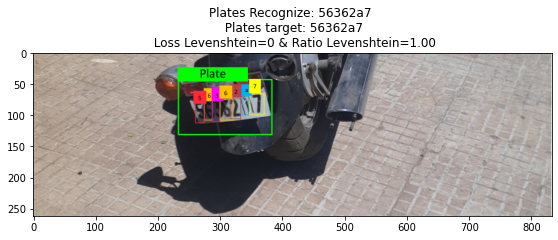}
    \caption{Plates Recognize: 56362a7, Plates target: 56362a7, Levenshtein distance=0 \& Ratio Levenshtein=1.00}
   \label{fig_Ng1} 
\end{figure}
\begin{figure}[htbp]
    \includegraphics[scale=0.45]{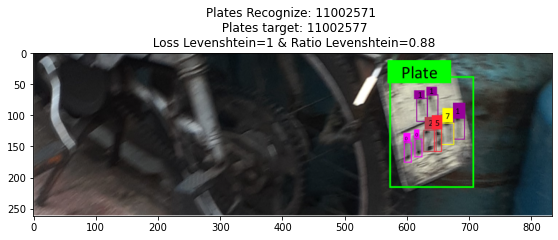}
    \caption{Plates Recognize: 11002571, Plates target: 11002577, Levenshtein distance =1 \& Ratio Levenshtein=0.88}
   \label{fig_Ng2}
\end{figure}

\begin{figure}[htbp]
   \includegraphics[scale=0.45]{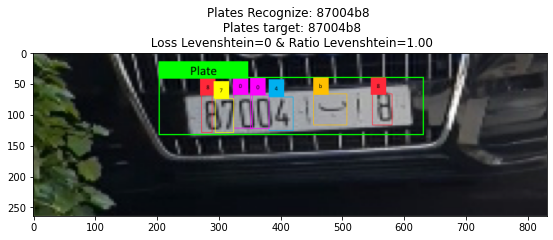}
    \caption{Plates Recognize: 87004b8, Plates target: 87004b8 , Levenshtein distance =0 \& Ratio Levenshtein=1.00}
   \label{fig_Ng3}
\end{figure}

\end{center}

\end{document}